# Stylistic Analysis of the French Presidential Speeches:
## Is Macron really different?

Dominique Labbé, Jacques Savoy[0000−0002−4486−0067]

| | |
|---|---|
| University of Grenoble | University of Neuchatel |
| Pacte/IEP - BP 48 | rue Emile Argand 11 |
| 38040 Grenoble cedex 9, France | 2000 Neuchatel, Switzerland |
| Dominique.Labbe@umrpacte.fr | Jacques.Savoy@unine.ch |

**Abstract**

Presidential speeches indicate the government's intentions and justifications supported by a dedicated style and rhetoric oscillating between explanation and controversy. Over a period of sixty years, can we observe stylistic variations by the different French presidents of the Fifth Republic (1958–2018)? Based on official transcripts of all their allocutions, this paper illustrates the stylistic evolution and presents the underlying main trends. This study shows that de Gaulle's rhetoric is not mainly dedicated to his own person, or that the two terms of J. Chirac are not fully similar. According to several overall stylistic indicators, Macron's style does not appear as complex compared to his predecessors (F. Hollande or N. Sarkozy) but a more careful analysis clearly demonstrates his noticeable new style. Compared to the recent US presidents, the French ones present some similarities (e.g., similar mean sentence length) and dissimilarities (more I-words, less we-words). In this comparative analysis, Macron's style is also clearly distinctive from both the US and former French presidents. Opting for a more abstract discourse, less anchored in space, using less numbers, E. Macron tends to use long sentences. These various stylistic and rhetorical features could explain his being misunderstood by the French people and his recurrent low approval ratings.

## 1 Introduction

The presidential speeches own different objectives such as explaining a situation, presenting several solutions, justifying governmental actions, understanding the pain of the people (e.g., after a terrorist attack or a natural disaster), convincing his electoral base of his own leadership or showing his strength and determination. Delivering such speeches constitutes the more visible daily activity of the French presidents.

Until recently, the number of presidential speeches remained low (e.g., 27 per year under Giscard d'Estaing presidency (1974–1981)), but reached an average close to one speech per day under Hollande's presidency (2012–2017). Such an increase could be explained by the growing importance of the press, including TV and the social networks. In addition, there is a clear motivation to speak directly to the public opinion instead of being limited to discussions with the parliament. Caesar *et al.* (1981) conclude that "speaking is governing" and in a recent newspaper article, Trump was called the "Narrator in Chief" (Parker & Costa, 2019).

To convince their fellow-citizens, presidents must adopt an effective style and rhetoric. In this study, rhetoric is defined as the art of effective and persuasive speaking, the way to motivate an audience, while an author's persuasive style is evaluated through studying frequent forms used in order to support his/her communication objective but also chosen for aesthetical value (Biber & Conrad 2009).

Having collected all of the speeches delivered by the French presidents of the fifth Republic, the studied corpus begins with C. de Gaulle in 1958 and ends with E. Macon in 2018. With this large dataset, can one extract the stylistic patterns characterizing each French president? Can we detect a few trends over time, for example, towards a more simple and direct speech, a style adopted by D. Trump during his electoral campaign (Savoy, 2018a; 2018b) or as president (Savoy, 2017)? Moreover, does the second term of a president differ from the first one?

This paper is organized as follows. The next section presents some related research in computer-aided analysis of political speeches. The third section describes our corpus and indicates some overall statistics. The fourth reports overall stylistic measurements and compares them across all ten French presidencies. The fifth exposes our characteristic vocabulary method based on occurrence frequencies and parts of speech. The sixth applies this method to the French presidential speeches and, in particular, to define Macron's rhetoric and style. A conclusion draws the main findings of this study.

## 2   State of the Art

According to Biber & Conrad (2009), a stylistic study should be based on ubiquitous and frequent words. As an operational definition, the stylometric analysis could be based on the most frequent word-types (MFWs) or lemmas, and values between 200 and 500 have been justified in author attribution studies (Burrows, 2002; Savoy, 2015a). When inspecting these MFWs, one can observe a large number of functional words (e.g., determiners, prepositions, conjunctions, pronouns, and auxiliary verb forms) used unconsciously by the author and without a clear and important meaning. These properties allow them to closely reflect a presidential style and thus can be pertinent features for establishing an intertextual distance between texts.

Based on such a distance, an automatic classification algorithm (Baayen, 2008; Jockers, 2014; Arnold & Tilton, 2015) can visualize similarities between texts (or samples of them) written by the same author or during the same time period. This allows analysts to draw maps showing the stylistic similarities between US presidents (Savoy, 2017) or to draw graphic affinities according to the main topics of their speeches (Savoy, 2015b). Other studies offer similar approaches to detect and monitor topics over time (Rule *et al.*, 2015). Instead of being limited to the MFWs, the entire vocabulary can be analyzed to compute an intertextual distance (Labbé, 2007), an approach that "preserves quite a rich body of information" (Love, 2002, p. 214).

Various studies have proposed to describe the style of each author by a few vocabulary richness indicators (Juola *et al*., 2019) such as Type-Token Ratio (TTR) (Hart, 1984), hapax density (percentage of words occurring once), or the lexical density (LD) (Biber *et al*., 2002) defined in Equation 1.

$$\text{LD}(t) = \text{lexical word}(t) / n(t) = 1 - \left(\text{function word}(t) / n(t)\right) \tag{1}$$

In this formulation, the variable *n(t)* indicates the total number of tokens (or the text length) of a text *t*, *function words(t)* the number of function words in *t*, *lexical word(t)* the number of lexical words in *t*. This latter set is composed of nouns, names, adjectives, verbs, and adverbs. On the other hand, function words regroup all other grammatical categories, namely determiners (e.g., *the*, *this*), pronouns (e.g., *you*, *us*), prepositions (e.g., *to*, *in*), conjunctions (e.g., *and*, *or*), modal verbs and auxiliary verb forms (e.g., *has*, *would*, *can*). A relatively high LD percentage indicates a more complex text, containing more information.

As another overall measurement, the mean word length, measured in number of letters, could also reflect the message's complexity (Hart, 1984; Tausczik & Pennebaker, 2010), the longer the mean word length, the higher the complexity. Instead of computing this mean word length, Tausczik & Pennebaker (2010) suggest to consider the percentage of Big Words (BW) defined as composed of six letters or more. Such a relationship between complexity and word length is clearly established:

> "One finding of cognitive science is that words have the most powerful effect on our minds when they are simple. The technical term is basic level. Basic-level words tend to be short. … Basic-level words are easily remembered; those messages will be best recalled that use basic-level language." (Lakoff & Wehling, 2012, p. 41)

For example, L. B. Johnson (presidency: 1963–1969) recognized the fear of having a too complex style by specifying to his ghostwriters: "I want four-letter words, and I want four sentences to the paragraph." (Sherrill, 1967).

In addition, a stylistic study may focus on syntactical aspects such as the mean sentence length (MSL) (measured by the average number of tokens). The occurrence of long sentences denotes a substantiated reasoning or indicates a detailed explanation. Even if a long sentence is required, its length is usually not conducive to an easy understanding.

Instead of analyzing all words or generating a single value as a stylistic indicator, various studies attempt to regroup several word-types under a semantic tag. Such approaches are however limited to the English language. For example, in the category *Symbolism,* Hart (1984), and Hart *et al*. (2013) include a list of words related to country (e.g., *nation*, *America*), ideology (e.g., *democracy, freedom, peace*), or generally, political concepts and institutions (e.g., *law*, *government*). The underlying hypothesis is to assume that the words serve as guides to the way the author thinks, acts, or feels. Based on US presidential speeches, Hart (1984) with his DICTION system describes the rhetorical and stylistic differences between the US presidents from Truman to Reagan. In a follow-up study, Hart *et al*. (2013) exposes the stylistic variations from G.W. Bush to Obama.

Based on a similar strategy, the system LIWC (Linguistic Inquiry & Word Count) (Tausczik & Pennebaker, 2010) regroups words under syntactical, emotional or psychological categories. Such classes may correspond to specific grammatical categories (e.g., first person singular denoted *Self* with *I, me, mine, my*), broader classes (e.g., personal pronouns) as well as more complex ones (verbs in the past tense, auxiliary verbs). With some semantics, the LIWC system defines positive emotions (*Posemo*) (e.g., *happy, hope, peace*) or negative (*Negemo*) (e.g., *humiliat\*, war*), *Cognition* (e.g., *admitted, perceive*) or terms related to *Human* (e.g., *family, friend, child\**). For example, using the LIWC system, Slatcher *et al*. (2007) were able to determine the personalities of different political candidates (US presidential election in 2004). They defined their psychological portraits both on single measurements (e.g., the relative frequency of different pronouns, positive emotions, etc.) and using a set of composite indices reflecting the cognitive complexity, presidentiality or honesty of each candidate. Automatically extracting political tone or psychological traits from texts is not exempt of concerns (Grimmer & Stewart, 2013).

To detect words or expressions overused or under-used by an author, Muller (1992) defines a measure of lexical characteristics. Focusing on governmental speeches written in French, Labbé & Monière (2003; 2008) have created a set of governmental corpora such as the *Speeches from the Throne* (Canada & Quebec), a corpus of the general policy statements of French governments (Labbé & Monière, 2003; 2008) as well as a collection of press releases covering the French presidential campaign of 2012 (Labbé & Monière, 2013), (Arnold & Labbé, 2015). Similar researches had been conducted with other languages, such as Italian (Pauli & Tuzzi, 2009). From these analyses, one can observe, for example, that governmental institutions tend to smooth out the differences between political parties when exercising power. Moreover, the temporal period of the documents constitutes an important factor in explaining the variations between presidents or prime ministers. The presence of a strong leader is usually accompanied with a real change in the style and vocabulary of governmental speeches (Labbé & Monière, 2003) as well as exceptional events (e.g., worldwide war, deep economic depression). Focusing on the United States, recent studies confirm these findings, as for example, using the *State of the Union* addresses (Savoy, 2015b) or inaugural addresses (Kubát & Cech, 2016).

## 3  Corpus

The messages and transcripts of the presidential allocutions have been downloaded from the Elysée website (www.elysee.fr) or from the website *Vie Publique* (www.vie-publique.fr). For each speech, a careful preprocessing has been done to obtain a single entry for possible spelling variations (e.g., Abu Dabi, Abou Dhabi, Abou-Dabi, etc.).

The second task was to assign to each word its Part-Of-Speech (POS) tag and corresponding lemma (headword, entry in the dictionary). As the English language has a relatively simple morphology, working on inflected forms (e.g., *I*, *me*, *mine*, or *laws*, *law*) or lemmas (such as *I* or *law*) often leads to similar conclusions. If the definition of a lemma is clear, the term "word" is ambiguous. The expression word-token (or simply token) refers to an occurrence or instance of a word-type (or type). For example, the sentence "The law is harsh but it is the law" counts nine tokens for six word-types (the, law, is, harsh, but, it), and six lemmas (the, law, (to) be, harsh, but, it).

In this study, the statistics are usually computed on lemmas. Taking account of lemmas instead of word-types implies slight variations in the count of the lexical items for the English language. With the French language however, each lemma can occur in several distinct tokens due to a richer morphology (Sproat, 1992).

To precisely define each lemma, a POS tagger (Manning & Schütze, 1999) must provide the correct grammatical category for each token. This operation could be done automatically but with errors, or semi-automatically with manual corrections. Our corpus was generated according to the second, time-consuming strategy producing an error-free tagged collection. As for the English, the French language presents ambiguities that cannot be removed without considering the semantics. For example, the sentence "Je suis le président" can be translated as "I'm the president" (said Macron) or as "I'm following the president" (said the bodyguard). The meaning is very different. As another example, one can consider the verbal form "est" corresponding to "is" or "east". Finally, the period sign does not always indicate the end of the sentence as in expressions such as "U.S.", "Mr. President", or in "Sept. 11th".

The function of the French presidency clearly presents strong similarities with the US one. In both cases, the president is the chief of the State, at the head of the administration, Commander-in-Chief, leader of a political party and at the head of the cabinet (partially in France, a function shared with the prime minister). Sharing a similar function, it is possible to compare the US and French presidential speeches, despite their cultural and language differences.

**Table 1.** Some Statistics About Presidents' Speeches and Comments

|  | **Presidency** | **Nb of texts** | **Texts/year** | **Tokens** | **Word-Types** |
|---|---|---|---|---|---|
| De Gaulle | 1958-1969 | 459 | 44.4 | 410,492 | 9,002 |
| Pompidou | 1969-1974 | 139 | 27.8 | 259,918 | 8,076 |
| Giscard d'Estaing | 1974-1981 | 191 | 27.3 | 660,560 | 9,535 |
| Mitterrand 1 | 1981-1987 | 1,590 | 227.1 | 3,363,824 | 20,861 |
| Mitterrand 2 | 1987-1995 | 957 | 136.7 | 2,212,420 | 18,635 |
| Chirac 1 | 1995-2002 | 1,305 | 186.4 | 2,444,858 | 19,834 |
| Chirac 2 | 2002-2007 | 1,173 | 234.6 | 1,636,839 | 16,160 |
| Sarkozy | 2007-2012 | 1,074 | 214.8 | 3,221,250 | 24,597 |
| Hollande | 2012-2017 | 1,544 | 308.8 | 3,182,939 | 20,403 |
| Macron | 2017-2018 | 318 | 200.8 | 1,037,988 | 14,819 |
| **Total** | 1958-2018 | 8,748 |  | 18,431,088 | 45,575 |

Table 1 reports the ten French presidencies of the fifth Republic with their president's name, year in office, the number of speeches (under the row "Nb of texts"), the number of speeches per year ("Texts/year"), their length ("Tokens"), and the vocabulary size (number of distinct word-types). The last row indicates the length of the entire corpus (Arnold *et al.*, 2016). Compared to Melville's *Moby Dick,* containing around 250,000 tokens, the size of this collection represents 73 times this novel.

As with the US presidents, one can infer that the number of speeches per year is increasing, reaching 227 during the Mitterrand's first term, 215 with Sarkozy's presidency or 309 with Hollande. With Macron, this value tends to decrease to 200.8 (= 318 / 1,6 year) depicting a first distinction with his two direct predecessors. Compared to the US presidents, the mean number of speeches per year is also smaller than the number of allocutions delivered by Obama on an annual basis (e.g., 491 in 2010).

This corpus with its tagged and checked version is freely available (www.unine.ch/clc/). The most pertinent aspect of this collection is the presence of the POS categories and the lemmas allowing the user to conduct various researches in contemporary French written language.

## 4  Overall Measurements

One can consider that speeches delivered by the presidents correspond also to an oral communication form. However, as mentioned by Biber & Conrad (2009, p. 262):

> "Language that has its source in writing but performed in speech does not necessarily follow the generalization [written vs. oral]. That is, a person reading a written text aloud will produce speech that has the linguistic characteristics of the written text. Similarly, written texts can be memorized and then spoken".

Thus, when comparing overall stylistic measurements, the reference must be with written texts and not oral dialogues. Table 2 depicts some overall stylistic measurements. In the second column, the type-token ratio (TTR) (Popescu *et al.*, 2009) is provided. A high value specifies the presence of a rich vocabulary showing that the underlying text covers many different topics or that the author presents a theme from several points of view with different formulations. To compute this value, one divides the vocabulary size (number of word-types) by the text length (number of tokens). This estimator has the drawback of being unstable, tending to decrease with text length (Baayen 2008). To avoid this problem, the computation could be based on a moving average (Popescu *et al.*, 2009; Covington & McFall, 2010). In this study, these values have been computed as an average value per 10,000 non-overlapping tokens. Thus, the first segment begins with the first word to the word appearing at position 10,000 while the second chunk starts at position 10,001 to 20,000, etc. This strategy is applied to avoid comparing texts with different lengths, a feature rendering the direct comparison between various measurements difficult.

In the remaining columns, the lexical density (or LD), the percentage of big words (BW), the average word length ("Word length"), and the mean sentence length (MSL) are reported. For each column, the maximal values are depicted in bold.

The values depicted in Table 2 correspond more to a written message with LD values higher than 0.4 (Biber *et al.*, 2002) and MSL larger than 20. The difference between the nine remaining French presidencies is rather small. De Gaulle possesses the highest TTR while Giscard d'Estaing has the lowest, indicating, for the latter, the presence of a less rich vocabulary and more repetitions of the same words and expressions. The Chirac's presidencies appear with the highest percentage of BW, indicating the presence of more complex allocutions. His high mean word length tends to confirm this finding. Over the years, one can detect a trend to a shorter MSL signaling certainly a concern to be better understood by the audience and the press.

**Table 2.** Overall Stylistic Measurements for the Ten French Presidencies (out of 10,000 word-tokens)

| President | TTR | LD | BW | Word Length | MSL |
|---|---|---|---|---|---|
| De Gaulle | **23.57**% | 49.06% | 32.42% | 4.66 | 30.75 |
| Pompidou | 22.43% | 49.78% | 31.53% | 4.62 | 30.27 |
| Giscard d'Estaing | 19.79% | 49.65% | 31.89% | 4.65 | 26.57 |
| Mitterrand 1 | 22.61% | 50.22% | 31.06% | 4.60 | 27.00 |
| Mitterrand 2 | 22.73% | 50.52% | 31.02% | 4.59 | 24.75 |
| Chirac 1 | 22.56% | 50.68% | 33.53% | 4.77 | 24.10 |
| Chirac 2 | 22.34% | 50.24% | **33.91**% | **4.80** | 24.06 |
| Sarkozy | 21.04% | 50.59% | 30.89% | 4.58 | 22.71 |
| Hollande | 20.48% | **50.83**% | 31.64% | 4.65 | 24.45 |
| Macron | 20.87% | 49.77% | 32.50% | 4.73 | **33.72** |

Comparing Macron's style with his two direct predecessors (Sarkozy & Hollande), Table 2 indicates a higher percentage of BW, as well as a higher mean word length. But the most distinct stylistic feature appearing in Table 2 is the mean number of tokens per sentence (MSL) in Macron's speeches, reaching 33.7, the largest value over all presidencies. In addition, during half the time, he is using sentences of 42 tokens or more.

Compared to the last two US presidents, both the lexical density (LD) and percentage of big words (BW) are higher for the French presidents (Obama: LD: 46.4%; BW: 26.3%; Trump: LD: 47.6, BW: 29.4%) (Savoy, 2017). Both Obama and Trump exhibit a lower informative speech (lower LD) but delivered in a simpler form (lower BW). With an MSL of 20.2 for Clinton, 18.7 for Obama, or 18.5 for Trump, Sarkozy and Hollande appear similar to these three contemporary US presidents.

## 5 Characteristic Vocabulary

All presidents are speaking with similar words or expressions but the differences between them reside in their frequencies. To determine the terms overused (or under-used) by an author, Muller (1992) suggests analyzing the number of term occurrences between a writer compared to the whole corpus. For example, knowing that the name *France* appears 2,897 times in Macron's speeches compared to 75,493 occurrences in the entire corpus, can we infer that *France* is overused (C+), under-used (C-) or employed with a non-different frequency (C=) than the other French presidents?

To provide an answer to this question, the size (number of tokens) of the entire corpus is indicated by the variable $n_0$ (= 18,413,088, Table 1) while the sample by Macron is denoted $n_1$ (= 1,038,889). For the selected term (e.g., *France*), one can count its number of occurrences in the entire corpus (value denoted *tf₀*, or 75,493) and its absolute frequency in Macron's texts (denoted by *tf₁*, or 2,897). Assuming that the frequency for this term (denoted *t*) in Macron's sample is the same as for the other presidents (hypothesis $H_0$), one can estimate the probability that *t* = *France* appears *tf₁* times in Macron's speeches according to a hypergeometric law (Baayen, 2001 & 2008) described by the following equation:

$$p(t = tf_1) = \frac{\binom{tf_0}{tf_1} \cdot \binom{n_0 - tf_0}{n_1 - tf_1}}{\binom{n_0}{n_1}} \tag{2}$$

Estimating only the probability of a single occurrence number is not fully pertinent (and would be very small). One can admit that some natural variability does exist; one speaker might employ this word a little bit more or less, without depicting a statistically significant deviation. To define the thresholds determining the limits of this variability, one can compute the cumulative distribution of the occurrence frequency of the term *t* in the underlying corpus as indicated in Equation 3.

$$C_t = p(t \leq tf_1) = \sum_{f=0}^{f=tf_1} p(t = f) \tag{3}$$

where *f* represents the absolute frequency (*f* = 0, 1, 2, … *tf₁*) of the chosen word *t* in Macron's sample. The maximal frequency is *tf₀*, its number of occurrences in the whole corpus. To apply the statistical test, one can define the lower and upper frequency limits (confidence interval) for which the cumulative probability distribution reaches a specified threshold denoted α and 1-α (e.g., α = 5%, 2.5%, 1%, or 0.5%).

For example, according to the null hypothesis $H_0$ and specifying α = 5%, one can observe between 4,135 and 4,383 occurrences of *France* in Macron's speeches (assuming that this word-type is used with the same intensity as under the other presidencies). According to this model, one can expect observing, in mean, $n_1/n_2 \cdot tf_0$ times the term *France* in Macron's speeches (or 1,038,899 / 18,413,088 · 75,493 = 4,258). However, *France* occurs only 2,897 times in Macron's allocutions, too few occurrences to be inside the confidence interval. Thus, Macron employed the name *France* less often (and this name appears in his C- vocabulary).

Instead of using Equation 2 (a model based on a single urn), we propose to take account of the proportion of each of the nine main POS categories, namely: nouns, names, verbs, adjectives, adverbs, prepositions, conjunctions, determiners, and pronouns. Thus, the different occurrences are no longer drawn from the entire corpus (or $n_0$) but from the number of words belonging to the corresponding POS category in the entire corpus ($nc_0$) or in the target subset ($nc_1$). This proposed model now considers nine distinct urns, one per POS category.

$$p(t = tf_1) = \frac{\binom{tf_0}{tf_1} \cdot \binom{nc_0 - tf_0}{nc_1 - tf_1}}{\binom{nc_0}{nc_1}} \tag{4}$$

Using this technique, one can determine some names characterizing Macron's speeches such as *Europe*, *Paris*, *Afrique*, *European Union*, *China*, *Sahel*, *Syria*, *Libya*, *Italy*, all belonging in C+ while *France*, *French* (people), *Germany*, or *United States* are under-used and appear in C- (in our examples, α was fixed to 1%).

## 6  Characteristic Vocabulary of the French Presidents

Focusing on names, one can extract the first interesting view of the vocabulary of the French presidents and the distinctive usage done by Macron. Table 3 shows the top ten most frequently occurring names by President Macron (under the label "Word-Type") with the corresponding rank in Macron's speeches (first column), and the rank obtained when considering all other presidents ("Rank Presidents). The column "Frequency" reports the relative frequency in ‰ in Macron's allocutions and the last column the difference (in percentage) between Macron and the nine other presidencies. As depicted in the first row, *France* is the most frequently used term for all presidents, achieving a relative frequency of 2.79‰ with Macron but showing a decrease of 32% compared to the other presidents.

Table 3 indicates the recurrent strategic interests of the French governments that are *France*, *Europe*, and *Africa*. These three targets do not change under Macron's presidency. With Macron however, *Sahel* appears high in this list due to a war in which the French troops are engaged. *China* receives more attention by Macron (+68.8%), but both *Germany* and *United States* receive less interest. With the decrease in frequency of the names *France* (-32%) and *French* (people) (-39.4%) on the one hand, and on the other the increase of *Europe* (4.4%) and more significantly *European Union* (21.4%), Macron seems to be a president less concerned (compared to the previous presidents) by his own country than by Europe.

Table 3. The Top Ten Most Frequently Used Names by E. Macron

| Rank Macron | Rank Presidents | Word-Type | Frequency | Difference |
|---|---|---|---|---|
| 1 | 1 | France | 2.79‰ | -32.0% |
| 2 | 2 | Europe | 1.91‰ | 4.4% |
| 3 | 3 | French (people) | 0.57‰ | -39.4% |
| 4 | 4 | Paris | 0.52‰ | 30.0% |
| 5 | 5 | Africa | 0.37‰ | -5.1% |
| 6 | 7 | European Union | 0.34‰ | 21.4% |
| 7 | 87 | Sahel | 0.27‰ | 831.0% |
| 8 | 11 | China | 0.27‰ | 68.8% |
| 9 | 6 | Germany | 0.19‰ | -36.7% |
| 10 | 8 | United States | 0.18‰ | -35.7% |

Another interesting aspect of governmental speeches are the frequency differences in the usage of pronouns. Pennebaker (2011) reports that just after his election, the media asserted that Obama used the pronoun *I* too often. But on a comparative basis, Obama's 10.5‰ of *I*-words is clearly lower than the 12.6‰ of Clinton, or the 16.1‰ of G.H. Bush (the father). What characterized the US presidencies is a high frequency of *we*-words (we, us, our) (46.6‰ with Trump), depicting even a slightly higher frequency that the determiner *the* in Trump's speeches (Savoy, 2017).

For the French presidencies, Table 4 provides the relative frequencies of four personal pronouns. As a distinction exists between *you* in singular and plural in the French language, Table 4 indicates only the plural form which is also the politest expression. The frequency of you (singular, or "*tu*") is very small. The pronouns overused under a given presidency are depicted in italics (e.g., the *I*-words under both Mitterrand's terms).

**Table 4.** Relative Frequencies of the Personal Pronouns (in ‰)

|  | we (nous) | I (je) | s/he/it (il) | you (vous) |
|---|---|---|---|---|
| De Gaulle | 8.70‰ | 8.84‰ | *13.48‰* | 5.28‰ |
| Pompidou | 8.17‰ | 15.96‰ | *15.27‰* | 4.52‰ |
| Giscard d'Estaing | 7.19‰ | *17.33‰* | 15.86‰ | 6.76‰ |
| Mitterrand 1 | 8.06‰ | *21.52‰* | 12.57‰ | *8.22‰* |
| Mitterrand 2 | 8.01‰ | *19.76‰* | 13.64‰ | *7.89‰* |
| Chirac 1 | 8.14‰ | 14.39‰ | *11.16‰* | *6.05‰* |
| Chirac 2 | *9.06‰* | 12.84‰ | 10.18‰ | 5.46‰ |
| Sarkozy | *9.21‰* | 18.26‰ | 11.69‰ | *7.13‰* |
| Hollande | *13.75‰* | 12.58‰ | 14.96‰ | *5.62‰* |
| Macron | *15.54‰* | 14.03‰ | 9.59‰ | 6.21‰ |

Overall, the French presidents employ the *we*-words less often than the US presidents. With Sarkozy, Hollande and Macron, this pronoun is overused, but the relative frequencies reported in Table 4 are lower than those found with the recent US presidents (e.g., Obama: 39.6‰; Trump: 46.6‰). For a political leader, the *we*-words own the advantage of being ambiguous. What is behind a *we*? The president and the cabinet, the president and the Parliament? Often it is a form including the listeners, to create a direct contact with the people as in "you and me", together.

Another distinction between the French and US presidencies are the more frequent usage of *I*-words in France. The frequencies depicted under Giscard d'Estaing and both of Mitterrand's terms are clearly higher (C+) compared to US presidents (G. Ford exhibits the highest frequency over all 45 presidents with 17.1‰, followed by G.H. Bush: 16.1‰, L.B. Johnson: 13.5‰, and B. Clinton: 12.6‰). During a US electoral campaign however, the winning candidates employ *I*-words very frequently and less *we*-words (Savoy, 2018).

During De Gaulle's presidency, the press tended to view this president as a strong one, centered into himself with a high frequency of *I*-words. Table 4 depicts an opposite picture, and a possible explanation is to compare De Gaulle with presidents of the fourth French Republic, political figures that did not have the power attributed to the presidency under the fifth Republic.

The first three presidents in Table 4 share in common the overuse of the pronoun *s/he/it* strongly related to reality and, as a corollary, a speech more geared towards explanation. With the French presidents, the two terms of Mitterrand are similar with an overuse of the pronouns *I* and *you* (plural). This pattern corresponds to a rhetorical form which uses interpellation as a means of achieving a polemic aim. For Chirac, there is a significant difference between the first term with an overuse of *s/he/it* and *you* (plural), but only *we*-words in the second, the turning point was after his stroke (September 2005). This stylistic aspect is present in all of his successors, resulting in a less personal tone.

Finally, one can consider the entire vocabulary used by the French presidents and extract the few nouns, adjectives or verbs characterizing them. Table 5 reports the top five most overused lemmas (mainly nouns, adjectives, and verbs, all belonging to the C+ vocabulary) according to our characteristic vocabulary model.

**Table 5.** Examples of Overused Lemmas for each Presidency

|  | 1$^{st}$ lemma | 2$^{nd}$ lemma | 3$^{rd}$ lemma | 4$^{th}$ lemma | 5$^{th}$ lemma |
|---|---|---|---|---|---|
| De Gaulle | honor | Algeria | universe | Algerian | people* |
| Pompidou | Nixon | consequence | Heath | Parisian | highway |
| Giscard d'Estaing | currently | hour | detente | new | increase |
| Mitterrand 1 | I / me | good | few | new | talk |
| Mitterrand 2 | Maastricht | community | Gorbatchev | CSCE | I |
| Chirac 1 | naturally | our | Union | peace | european |
| Chirac 2 | the | international | development | Evian | Nepad |
| Sarkozy | not | crisis | will | euro | because |
| Hollande | also | power | we | here | Mali |
| Macron | we / our | exactly | transformation | commitment | and |

* "peuple" (singular) not "gens" (plural).

The data depicted in Table 5 confirms some of our previous findings and reveals others. For example, the *I*-words are in the first rank under the first Mitterrand presidency, but decreases during his 2$^{nd}$ term. The *we*-words appear in the first position under Macron's presidency, as well as the conjunction *and* (et), the term used to connect clauses together and generate long sentences.

External events and the political situation explain other lemmas occurring in Table 5. One of the most recurrent problem under De Gaulle's first term was the independence war in *Algeria*, the *detente* between East and West under Giscard, the European reconstruction (with *Maastricht* with Mitterrand 2, *European Union* with Chirac 1, *euro* with Sarkozy). In this table, CSCE is the acronym for "Conférence sur la Sécurité et la Coopération en Europe". *Africa* appears with other lemmas such as *Mali*, *Algeria*, or *NEPAD* (New Partnership for Africa's Development). A few personal names also appear in Table 5, such as President Nixon, English Prime Minister Heath, or President Gorbatchev.

## 7 Conclusion

Based on a large corpus of French presidential speeches, this study compares some stylistic measurements between the different French presidencies. Looking at the volume of speeches (see Table 1), one can observe a clear increase of the number of allocutions per year over the last three decades, reaching, in average, one address per day under Hollande's presidency (2012–2017).

Regarding the complexity of the addresses, one can see a trend towards a smaller mean sentence length (MSL) over the years (see Table 2). With De Gaulle (1958–1969), the MSL starts with a high value of 30.75 but reaches a minimal value of 22.7 under Sarkozy's presidency (2007–2012). A similar pattern can be detected with the Type-Toke Ratio (TTR), beginning with the largest value under De Gaulle (23.57%) and presenting the smallest value with Hollande (2012–2017). However, the lexical density (LD) or the percentage of big words (BW, words containing six letters of more) tends to be relatively stable across the ten presidencies. For this last measure, one can mention that both Chirac's presidencies display the highest values, depicting a distinct stylistic feature compared to the other French presidencies.

When analyzing the pronoun density, one can perceive clear individual differences between each president (see Table 4). The last two presidents (Hollande & Macron) use more frequently "we" while Mitterrand (1981–1995) presents a high frequency of "I/me". With De Gaulle, the density of pronouns is the lowest. On the other hand, the French presidential speeches expose also some constants. Looking at the most frequently used names over the lest 60 years (see Table 3), the main strategic topics stay the same, namely France, Europe, and Africa and in that order.

Compared to the US presidents, the LD of the French allocutions is higher indicating more informative speeches. On the other hand, the percentage of BW is also higher for the French presidents signaling allocutions that are more complex and not as easily understood by the listeners. When analyzing the MSL, the French governmental speeches contain a larger number of tokens (mean difference around 2) compared to the recent US presidents. For this variable, Macron depicts an atypically high average value of 34.7 tokens/sentence compared to 20.7 for Trump or 23.5 for Hollande-Sarkozy.

When focusing on some personal pronoun frequencies, the French presidents employ the *I*-words (I, me, mine) more often than the US presidents. The latter however use the *we*-words (we, us, ours) four times more frequently, striving to establish a dialogue with the public opinion. Between the French presidents, Macron exhibits a clear increase in the use of *we*-words, but his density is still three times lower than Trump's usage.

Compared to his two direct predecessors (Sarkozy & Hollande), Macron clearly uses longer sentences and his vocabulary includes more BW (see Table 2). Even if he shares the three main strategic interests with all his predecessors (France, Europe, Africa), he appears less concerned by his country and his fellow-citizens, but uses more often the name "Europe". He adopted a style tending to establish a fusion with the audience by using more *we*-words. Even if the French media call him Jupiter, our data cannot confirm a higher usage of the *I*-words under his presidency, reflecting a self-confident or arrogant political leader (Pennebaker, 2011).

Macron's presidency clearly presents a communication problem with sentences too long to be clearly understood by the public, built with many *and*s, generating a stream hard to follow for the listener (Smith, 2018). In addition, Macron's attitude to avoid press conferences, and especially to avoid questions from journalists, increases the absence of an understandable explanation or justification from the head of the government (Van Renterghem, 2018; Stangler, 2019).

**Acknowledgments**

This corpus of the French presidential speeches with its tagged and checked version is freely available (www.unine.ch/clc/). The authors would like to thank the anonymous referees for their helpful suggestions and remarks.